\documentclass[runningheads]{llncs}
\usepackage{graphicx}
\usepackage{amsmath}
\usepackage{amssymb}
\usepackage{booktabs}
\usepackage{multirow}
\usepackage{wrapfig}
\usepackage[table]{xcolor}

\usepackage{tikz}
\usepackage{comment}
\usepackage{amsmath,amssymb} 
\usepackage{color}
\usepackage{xspace}
\usepackage{hhline}
\usepackage[pagebackref,breaklinks,colorlinks,citecolor=blue,linkcolor=blue,bookmarks=false]{hyperref}
\usepackage{cite}
\usepackage{microtype}
\usepackage[accsupp]{axessibility}  
\usepackage{enumitem}
\usepackage[normalem]{ulem}
\usepackage[width=122mm,left=12mm,paperwidth=146mm,height=193mm,top=12mm,paperheight=217mm]{geometry}
\usepackage{xcolor, colortbl}
\usepackage{booktabs}
\usepackage[flushmargin]{footmisc}

\let\llncssubparagraph\subparagraph
\let\subparagraph\paragraph
\usepackage[compact]{titlesec}
\let\subparagraph\llncssubparagraph
\titlespacing{\section}{0pt}{3ex plus 1ex minus .2ex}{2ex minus .2ex}
\titlespacing{\subsection}{0pt}{2ex plus 1ex minus .2ex}{1ex minus .2ex}

\definecolor{darkblue}{rgb}{0, 0.2, 0.6}
\definecolor{orange}{rgb}{1.0, 0.5, 0.0}
\definecolor{red}{rgb}{1.0, 0.0, 0.0}
\definecolor{purple}{rgb}{0.6, 0.0, 0.6}
\definecolor{dogwoodrose}{rgb}{0.84, 0.09, 0.41}
\definecolor{mint}{rgb}{0.01176, 0.5490, 0.5490}
\definecolor{blue}{rgb}{0, 0, 1.0}
\definecolor{azure(colorwheel)}{rgb}{0.0, 0.5, 1.0}
\definecolor{nicegreen}{rgb}{0.0, 0.7, 0.1}
\definecolor{CuGray}{gray}{0.9}
\definecolor{amethyst}{rgb}{0.6, 0.4, 0.8}
\definecolor{black}{rgb}{0.0, 0.0, 0.0}
\definecolor{steelblue}{rgb}{0.27, 0.51, 0.7}
\definecolor{brightcerulean}{rgb}{0.11, 0.67, 0.84}

\newcommand{\colorref}[1]{{\color{blue}{#1}}}

\makeatletter
\DeclareRobustCommand\onedot{\futurelet\@let@token\@onedot}
\def\@onedot{\ifx\@let@token.\else.\null\fi\xspace}
\def\eg{\emph{e.g}\onedot} 
\def\ie{\emph{i.e}\onedot}

\def\etal{\emph{et al}\onedot}
\makeatother

\setlist[itemize]{align=parleft,left=0pt}

\begin{document}
\pagestyle{headings}
\mainmatter
\def\ECCVSubNumber{0000}  

\title{
Cross-Attention of Disentangled Modalities for 3D Human Mesh Recovery with Transformers
} 
\titlerunning{FastMETRO} 
\authorrunning{J. Cho et al.} 
\author{
Junhyeong Cho${}^1$
\qquad
Kim Youwang${}^2$
\qquad
Tae-Hyun Oh${}^{2,3,}$\thanks{Joint affiliated with Yonsei University, Korea.}
}
\institute{
${}^1$Department of CSE 
\qquad
${}^2$Department of EE 
\qquad
${}^3$Graduate School of AI\\
Pohang University of Science and Technology (POSTECH), Korea \\
\email{\{junhyeong99, youwang.kim, taehyun\}@postech.ac.kr} \\
\url{https://github.com/postech-ami/FastMETRO}
}

\maketitle
\begin{abstract}
\vspace{-2.5mm}
Transformer encoder architectures have recently achieved state-of-the-art results on monocular 3D human mesh reconstruction, but they require a substantial number of parameters and expensive computations.
Due to the large memory overhead and slow inference speed, it is difficult to deploy such models for practical use.
In this paper, we propose a novel transformer encoder-decoder architecture for
3D human mesh reconstruction from a single image, called \mbox{\textit{FastMETRO}}. 
We identify the performance bottleneck in the \mbox{encoder-based} transformers 
is caused by the token design which introduces high complexity interactions among input tokens.
We disentangle the interactions via an \mbox{encoder-decoder} architecture,
which allows our model to demand much fewer parameters and shorter inference time.
In addition, we impose the prior knowledge of human body's morphological relationship 
via attention masking and mesh upsampling operations,
which leads to faster convergence with higher accuracy.
Our \mbox{FastMETRO} improves the \mbox{Pareto-front} of accuracy and efficiency, and clearly outperforms \mbox{image-based} methods on Human3.6M and 3DPW.
Furthermore, we validate its generalizability on FreiHAND.

\vspace{-1mm}
\vspace{-2mm}
\end{abstract}

\section{Introduction}
3D human pose and shape estimation models aim to estimate 3D coordinates of human body joints and mesh vertices.
These models can be deployed in a wide range of applications that require human behavior understanding, \eg, human motion analysis and human-computer interaction.
To utilize such models for practical use, 
monocular methods~\cite{bogo2016smplify,hmrKanazawa17,pavlakos_2018_mpve,kolotouros2019spin,jiang2020coherent,Kocabas_PARE_2021,sun2021monocular,dwivedi2021dsr,zhang2021pymaf,kolotouros2019cmr,Moon_2020_ECCV_I2L-MeshNet,lin2021metro,lin2021graphormer} estimate the 3D joints and vertices without using 3D scanners or stereo cameras.
This task is essentially challenging due to complex human body articulation, and becomes more difficult by occlusions and depth ambiguity in monocular settings.

\begin{figure}[t!]
    \centering
    \includegraphics[width=\columnwidth]{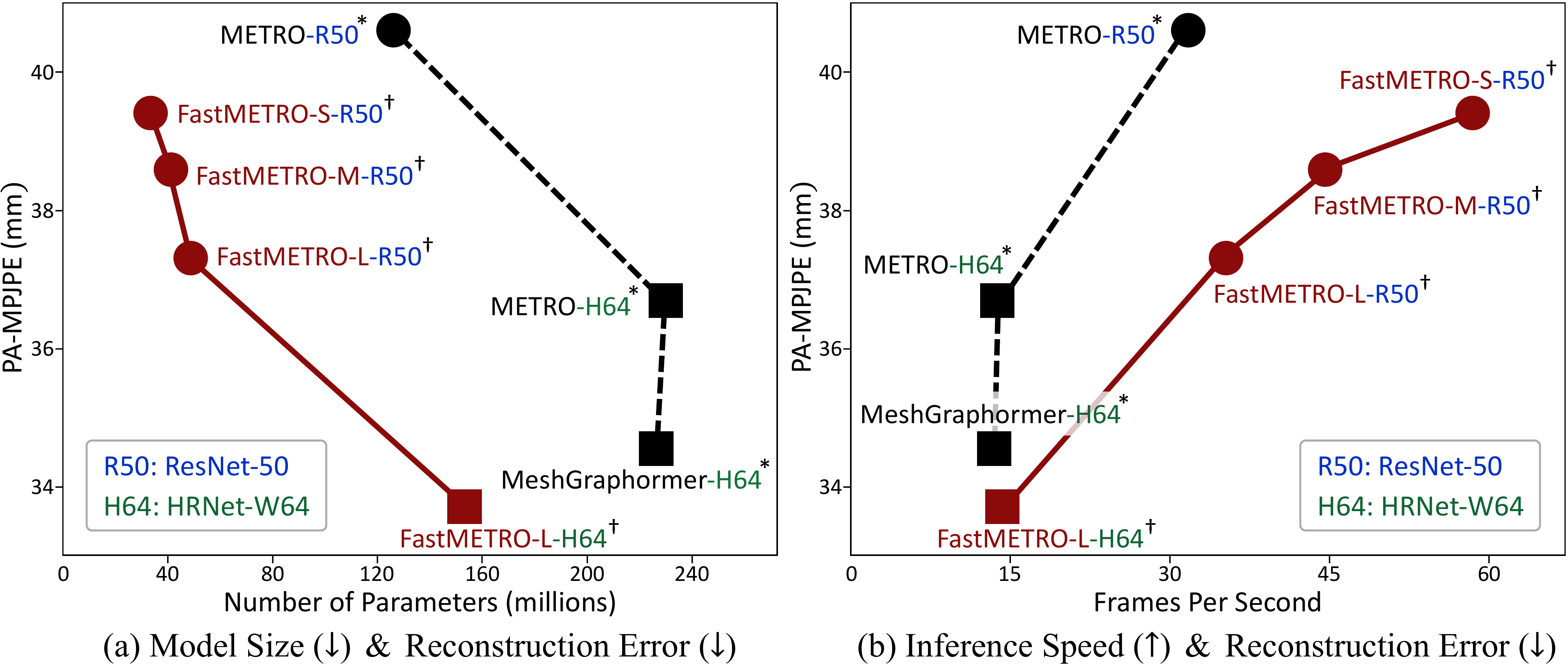}
    \caption{
    Comparison with encoder-based transformers~\cite{lin2021metro,lin2021graphormer} and 
    our models on Human3.6M~\cite{wang2014h36m}.
    Our FastMETRO substantially improves the Pareto-front of accuracy and efficiency.
    $\dagger$ indicates training for 60 epochs, and $\ast$ denotes training for 200 epochs.
    }
    \label{fig:comparison_params_fps}
\end{figure}

To deal with such challenges, state-of-the-art methods~\cite{lin2021metro,lin2021graphormer} exploit non-local relations among human body joints and mesh vertices via transformer encoder architectures.
This leads to impressive improvements in accuracy by consuming a substantial number of parameters and expensive computations as trade-offs; efficiency is less taken into account,
although it is crucial in practice.

In this paper, we propose \textbf{FastMETRO} which employs a novel transformer encoder-decoder architecture for 
3D human pose and shape estimation from an input image.
Compared with the transformer encoders~\cite{lin2021metro,lin2021graphormer}, \mbox{FastMETRO} is more practical because it
achieves competitive results with much fewer parameters and faster inference speed, as shown in Figure~\ref{fig:comparison_params_fps}.
Our architecture is motivated by the observation that the encoder-based methods overlook the importance of the token design which is a \mbox{key-factor} in accuracy and efficiency.

The encoder-based transformers~\cite{lin2021metro,lin2021graphormer} share similar transformer encoder architectures.
They take $K$ joint and $N$ vertex tokens as input for the estimation of 3D human body joints and mesh vertices, where $K$ and $N$ denote the number of joints and vertices in a 3D human mesh, respectively.
Each token is constructed by the concatenation of a global image feature vector $\mathbf x \in \mathbb R^{C}$
and 3D coordinates of a joint or vertex in the human mesh.
This results in the input tokens of dimension $\mathbb{R}^{(K+N)\times (C+3)}$ which are fed as input to the transformer encoders.\footnote{
For simplicity, we discuss the input tokens mainly based on METRO~\cite{lin2021metro}. 
Mesh Graphormer~\cite{lin2021graphormer} has subtle differences,
but the essence of the bottleneck is shared.
}
This token design introduces the same sources of the performance bottleneck: 1) spatial information is lost in the global image feature $\mathbf x$, and 2) the same image feature $\mathbf x$ is used in an overly-duplicated way.
The former is caused by the average pooling operation to obtain the global image feature $\mathbf x$.
The latter leads to considerable inefficiency, since expensive computations are required to process mostly duplicated information, where distinctively informative signals are only in $0.15\%$ of the input tokens.\footnote{
$3$-dimensional coordinates out of $(C+3)$-dimensional input tokens, where $C=2048$.
}
Furthermore, the computational complexity of each transformer layer is quadratic as $O(L^2C+LC^2)$, where $L \geq K+N$.
Once either $L$ or $C$ is dominantly larger, it results in unfavorable efficiency.
Both methods~\cite{lin2021metro,lin2021graphormer} are such undesirable cases.

\begin{figure}[t!]
    \centering
    \includegraphics[width=\columnwidth]{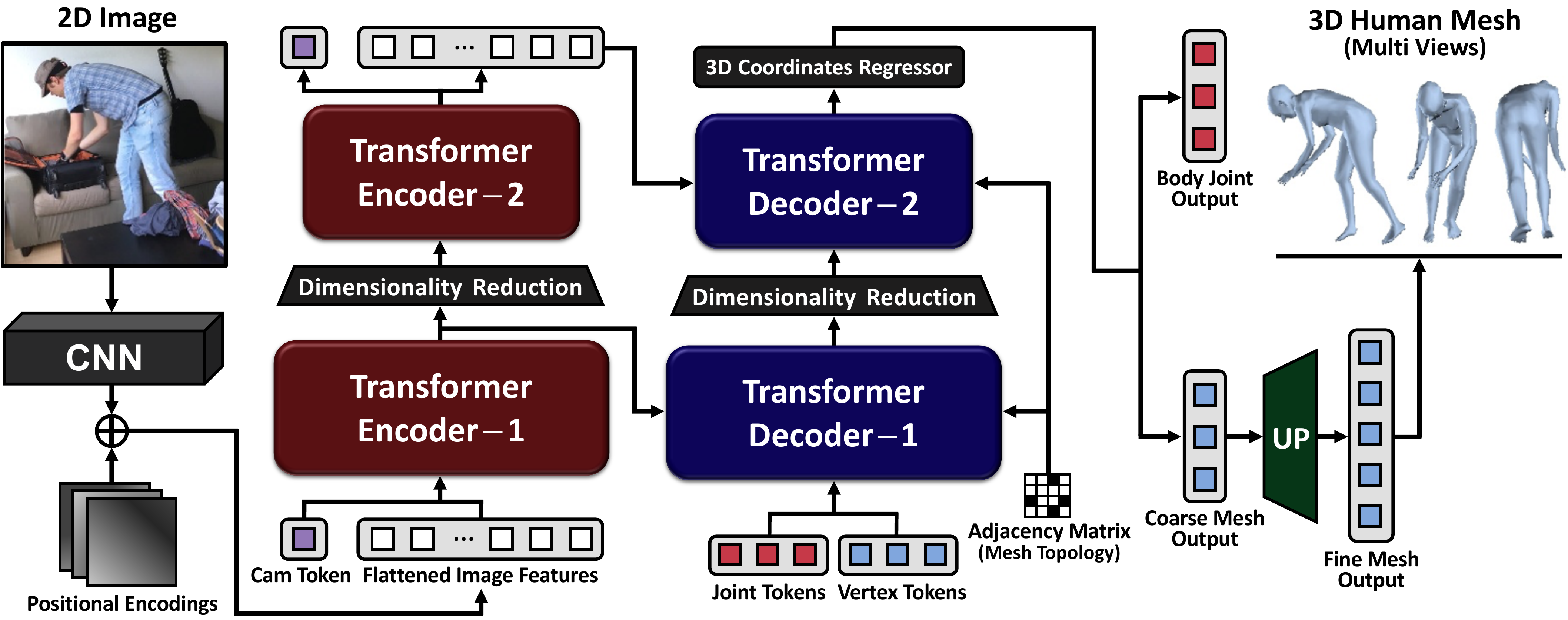}
    \caption{
    Overall architecture of FastMETRO.
    Our model estimates 
    3D coordinates of human body joints and mesh vertices from a single image.
    We extract image features via a CNN backbone, which are fed as input to our transformer encoder.
    In addition to image features produced by the encoder,
    our transformer decoder takes learnable joint and vertex tokens as input.
    To effectively learn \mbox{non-local} \mbox{joint-vertex} relations and local \mbox{vertex-vertex} relations, we
    mask \mbox{self-attentions} of \mbox{non-adjacent} vertices according to the topology of human triangle mesh.
    Following~\cite{lin2021metro,lin2021graphormer},
    we progressively reduce the hidden dimension sizes via linear projections
    in our transformer.
    }
    \label{fig:OverallArch}
\end{figure}

In contrast, our FastMETRO does not concatenate an image feature vector for the construction of input tokens.
As illustrated in Figure~\ref{fig:OverallArch},
we disentangle the image encoding part and mesh estimation part via an \mbox{encoder-decoder} architecture.
Our joint and vertex tokens focus on certain image regions through \mbox{cross-attention} modules in the transformer decoder.
In this way, the proposed method efficiently estimates the 3D coordinates of human body joints and mesh vertices from a 2D image.
To effectively capture non-local joint-vertex relations and local vertex-vertex relations,
we mask self-attentions of non-adjacent vertices according to the topology of human triangle mesh.
To avoid the redundancy caused by the spatial locality of human mesh vertices,
we perform \mbox{coarse-to-fine} mesh upsampling as in~\cite{kolotouros2019cmr,lin2021metro,lin2021graphormer}.
By leveraging the prior knowledge of human body's morphological relationship,
we substantially reduce optimization difficulty.
This leads to faster convergence with higher accuracy.

We present the proposed method with model-size variants by changing the number of transformer layers: FastMETRO-S, FastMETRO-M, \mbox{FastMETRO-L}. 
Compared with the encoder-based transformers~\cite{lin2021metro,lin2021graphormer}, 
\mbox{FastMETRO-S} requires only about 9\% of the parameters in the transformer architecture, but shows competitive results with much faster inference speed.
In addition, the large variant (\mbox{FastMETRO-L}) achieves the state of the art on the Human3.6M~\cite{wang2014h36m} and 3DPW~\cite{marcard20183dpw} datasets among image-based methods, which also demands fewer parameters and shorter inference time compared with the encoder-based methods.
We demonstrate the effectiveness of the proposed method by conducting extensive experiments, 
and validate its generalizability by showing 3D hand mesh reconstruction results on the FreiHAND~\cite{zimmermann2019freihand} dataset.

\noindent Our contributions are summarized as follows:
\begin{itemize}
  \item[$\bullet$] 
  We propose FastMETRO which employs a novel transformer encoder-decoder architecture for 3D human mesh recovery from a single image. 
  Our method resolves the performance bottleneck in the encoder-based transformers, and improves the Pareto-front of accuracy and efficiency.
  \item[$\bullet$]
  The proposed model converges much faster by reducing optimization difficulty.
  Our FastMETRO leverages the prior knowledge of human body's morphological relationship,
  \eg, masking attentions
  according to the human mesh topology.
  \item[$\bullet$]
  We present model-size variants of our FastMETRO.
  The small variant shows competitive results
  with much fewer parameters and faster inference speed.
  The large variant clearly outperforms existing image-based methods on the Human3.6M and 3DPW datasets, which is also more lightweight and faster.
\end{itemize}

\section{Related Work}
Our proposed method aims to estimate the 3D coordinates of human mesh vertices from an input image by leveraging the attention mechanism in the transformer architecture.
We briefly review relevant methods in this section.

\noindent \textbf{Human Mesh Reconstruction.}
The reconstruction methods belong to one of the two categories: parametric approach and non-parametric approach.
The parametric approach learns to estimate the parameters of a human body model such as SMPL~\cite{SMPL:2015}.
On the other hand, the non-parametric approach learns to directly regress the 3D coordinates of human mesh vertices.
They obtain the 3D coordinates of human body joints via
linear regression from the estimated mesh. 

The reconstruction methods in the parametric approach~\cite{guan2009estimating,bogo2016smplify,hmrKanazawa17,pavlakos_2018_mpve,kolotouros2019spin,jiang2020coherent,Kocabas_PARE_2021,sun2021monocular,dwivedi2021dsr,zhang2021pymaf} have shown stable
performance in monocular 3D human mesh recovery.
They have achieved the robustness to environment variations
by exploiting the human body prior encoded in a human body model such as SMPL~\cite{SMPL:2015}.
However, their regression targets 
are difficult for deep neural networks to learn;
the pose space in the human body model is expressed by the 3D rotations of human body joints, where
the regression of the 3D rotations is challenging~\cite{mahendran2018regression}.

Recent advances in deep neural networks have enabled the non-parametric approach with promising performance~\cite{kolotouros2019cmr,choi2020pose2mesh,Moon_2020_ECCV_I2L-MeshNet,lin2021metro,lin2021graphormer}.
Kolotouros \etal~\cite{kolotouros2019cmr} propose a graph convolutional neural network (GCNN)~\cite{kipf2017gcnn} to effectively learn local \mbox{vertex-vertex} relations, where the graph structure is based on the topology of SMPL human triangle mesh~\cite{SMPL:2015}.
They extract a global image feature vector through a CNN backbone, then construct vertex embeddings 
by concatenating the image feature vector with the 3D coordinates of vertices in the human mesh.
After iterative updates via graph convolutional layers, they 
estimate the 3D locations of human mesh vertices.
To improve the robustness to partial occlusions,
Lin~\etal~\cite{lin2021metro,lin2021graphormer}
propose transformer encoder architectures which effectively learn the non-local relations among human body joints and mesh vertices via the attention mechanism in the transformer.
Their models, METRO~\cite{lin2021metro} and Mesh Graphormer~\cite{lin2021graphormer}, follow the similar framework with the GCNN-based method~\cite{kolotouros2019cmr}.
They construct vertex tokens by attaching a global image feature vector to the 3D coordinates of vertices in the human mesh.
After several updates via transformer encoder layers, they regress the 3D coordinates of human mesh vertices.

Among the reconstruction methods, METRO~\cite{lin2021metro} and Mesh Graphormer~\cite{lin2021graphormer} are the most relevant work to our FastMETRO.
We found that the token design in those methods leads to a substantial number of unnecessary parameters and computations.
In their architectures, transformer encoders take
all the burdens to learn complex relations among mesh vertices,
along with the highly \mbox{non-linear} mapping between 2D space and 3D space.
To resolve this issue, we disentangle the image-encoding and mesh-estimation parts via an \mbox{encoder-decoder} architecture.
This makes FastMETRO more lightweight and faster, and allows our model to learn the complex relations more effectively.

\noindent \textbf{Transformers.}
Vaswani~\etal~\cite{vaswani2017attention} introduce a transformer architecture 
which effectively learns long-range relations through the attention mechanism in the transformer.
This architecture has achieved impressive improvements in diverse computer vision
tasks~\cite{carion2020end,guo2020normalized,dosovitskiy2021an,wang2021end,lu2021context,rao2021dynamicvit,liu2021paint,kim2021hotr,liu2021Swin,lin2021metro,lin2021graphormer,zheng2021poseformer,cho2021gsrtr,cho2022CoFormer}.
Dosovitskiy~\etal~\cite{dosovitskiy2021an} present a transformer encoder architecture, where a learnable token
aggregates image features via \mbox{self-attentions}
for image classification.
Carion~\etal~\cite{carion2020end} propose a transformer \mbox{encoder-decoder} architecture, where learnable tokens focus on certain image regions via \mbox{cross-attentions}
for object detection.
Those transformers have the most relevant architectures to our model.

Our FastMETRO employs a transformer encoder-decoder architecture, whose decoupled structure is favorable to learn the complex relations between the heterogeneous modalities of 2D image and 3D mesh.
Compared with the existing transformers~\cite{carion2020end,guo2020normalized,dosovitskiy2021an,wang2021end,lu2021context,rao2021dynamicvit,liu2021paint,kim2021hotr,zheng2021poseformer,cho2021gsrtr,cho2022CoFormer}, we progressively reduce hidden dimension sizes in the transformer architecture as in~\cite{lin2021metro,lin2021graphormer}.
Our separate decoder design enables FastMETRO to easily impose the human body prior by masking self-attentions of decoder input tokens, which leads to stable optimization and higher accuracy.
This is novel in transformer architectures.

\section{Method}
\begin{figure}[t!]
    \centering
    \includegraphics[width=\columnwidth]{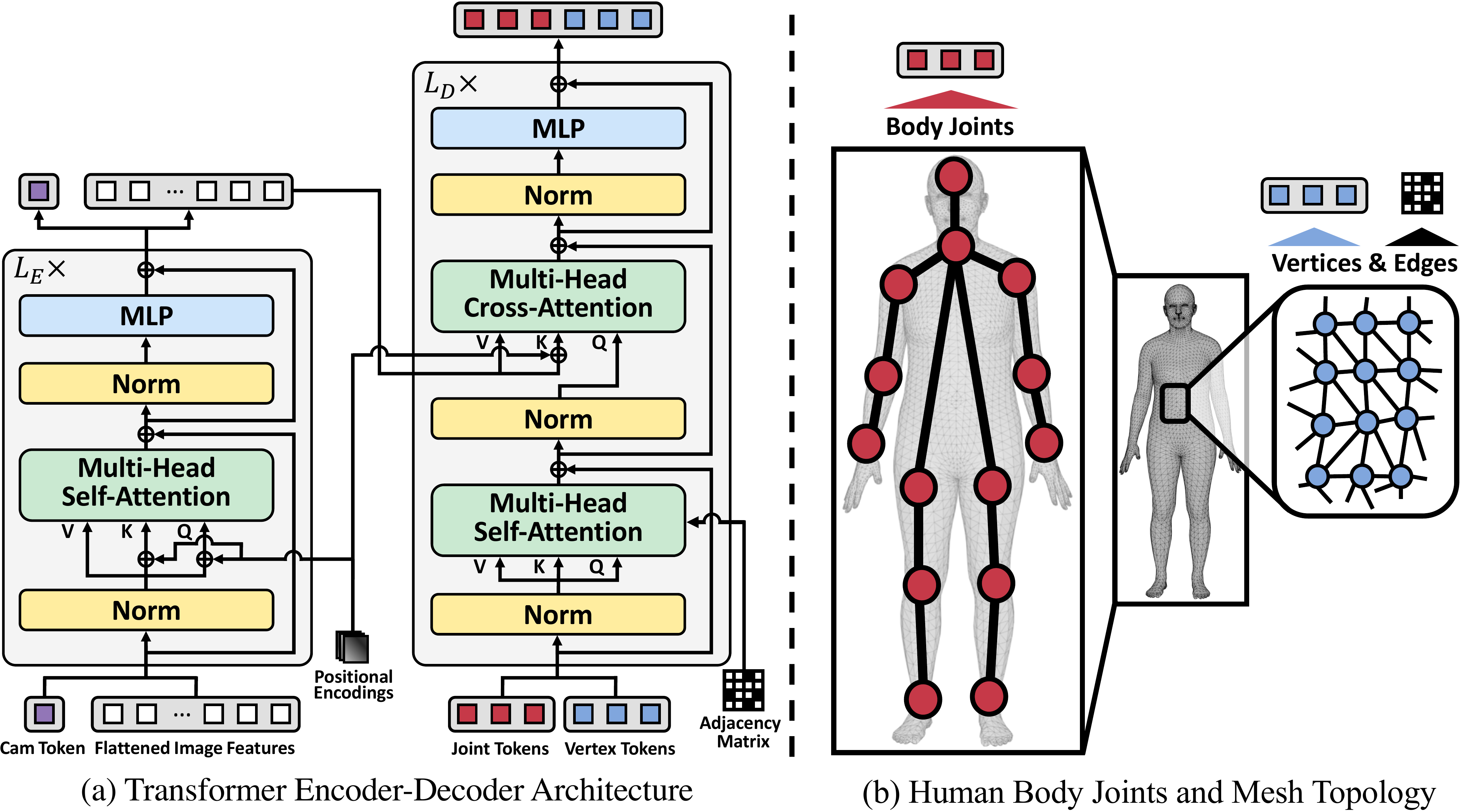}
    \caption{
    Details of our transformer architecture and 3D human body mesh.
    For simplicity, we illustrate the transformer without progressive dimensionality reduction.
    Note that the camera feature is not fed as input to the decoder.
    We mask attentions using the adjacency matrix obtained from the 
    human triangle mesh of SMPL~\cite{SMPL:2015}.
    }
    \label{fig:DetailArch}
\end{figure}

We propose a novel method, called \textbf{Fast} \textbf{ME}sh \textbf{TR}ansf\textbf{O}rmer (FastMETRO).
FastMETRO has a transformer encoder-decoder architecture 
for 3D human mesh recovery from an input image.
The overview of our method is shown in Figure~\ref{fig:OverallArch}.
The details of our transformer encoder and decoder are illustrated in Figure~\ref{fig:DetailArch}.

\subsection{Feature Extractor}
Given a single RGB image, our model extracts image features $\mathbf X_{I} \in \mathbb R^{H \times W \times C}$ through a CNN backbone, 
where $H \times W$ denotes the spatial dimension size and $C$ denotes the channel dimension size.
A $1 \times 1$ convolution layer takes the image features $\mathbf X_{I}$ as input, and reduces the channel dimension size to $D$.
Then, a flatten operation produces flattened image features $\mathbf X_{F} \in \mathbb R^{HW \times D}$.
Note that we employ positional encodings for retaining spatial information in our transformer, as illustrated in Figure~\ref{fig:DetailArch}.

\subsection{Transformer with Progressive Dimensionality Reduction}
Following the encoder-based transformers~\cite{lin2021metro,lin2021graphormer}, FastMETRO progressively reduces the hidden dimension sizes in the transformer architecture via linear projections, as illustrated in Figure~\ref{fig:OverallArch}.

\noindent \textbf{Transformer Encoder.}
Our transformer encoder (Figure~\ref{fig:DetailArch}\colorref{a}) takes a learnable camera token and the flattened image features $\mathbf X_{F}$ as input.
The camera token captures essential features to predict weak-perspective camera parameters through 
the attention mechanism in the transformer;
the camera parameters are used for fitting the 3D estimated human mesh to the 2D input image.
Given the camera token and image features,
the transformer encoder produces a camera feature and aggregated image features $\mathbf X_{A} \in \mathbb R^{HW \times D}$.

\noindent \textbf{Transformer Decoder.}
In addition to the image features $\mathbf X_{A}$ obtained from the encoder, our transformer decoder
(Figure~\ref{fig:DetailArch}\colorref{a})
takes the set of learnable joint tokens and the set of learnable vertex tokens as input.
Each token in the set of joint tokens $\mathcal{T}_{J} = \{\mathbf{t}^{J}_{1}, \mathbf{t}^{J}_{2}, \dots, \mathbf{t}^{J}_{K}\}$ is used to estimate 3D coordinates of a human body joint, where $\mathbf{t}^{J}_{i} \in \mathbb R^{D}$.
The joint tokens correspond to the body joints in Figure~\ref{fig:DetailArch}\colorref{b}. 
Each token in the set of vertex tokens $\mathcal{T}_{V} = \{\mathbf{t}^{V}_{1}, \mathbf{t}^{V}_{2}, \dots, \mathbf{t}^{V}_{N}\}$ is used to estimate 3D coordinates of a human mesh vertex, where $\mathbf{t}^{V}_{j} \in \mathbb R^{D}$.
The vertex tokens correspond to the mesh vertices in Figure~\ref{fig:DetailArch}\colorref{b}.
Given the image features and tokens, the transformer decoder produces joint features $\mathbf X_{J} \in \mathbb R^{K \times D}$ and vertex features $\mathbf X_{V} \in \mathbb R^{N \times D}$ 
through self-attention and cross-attention modules.
Our transformer decoder effectively captures non-local relations among human body joints and mesh vertices via \mbox{self-attentions}, which improves the robustness to environment variations such as occlusions.
Regarding the joint and vertex tokens, each focuses on its relevant image region
via \mbox{cross-attentions}.

\noindent \textbf{Attention Masking based on Mesh Topology.}
To effectively capture local vertex-vertex and non-local joint-vertex relations,
we mask self-attentions of non-adjacent vertices according to the topology of human triangle mesh in Figure~\ref{fig:DetailArch}\colorref{b}.
Although we mask the attentions of non-adjacent vertices, the coverage of each vertex token increases as it goes through decoder layers in the similar way with iterative graph convolutions.
Note that GraphCMR~\cite{kolotouros2019cmr} and Mesh Graphormer~\cite{lin2021graphormer} perform graph convolutions based on the human mesh topology, 
which demands additional learnable parameters and computations.

\subsection{Regressor and Mesh Upsampling}
\noindent \textbf{3D Coordinates Regressor.}
Our regressor takes the joint features $\mathbf X_{J}$ and vertex features $\mathbf X_{V}$ as input, and estimates the 3D coordinates of human body joints and mesh vertices.
As a result, 3D joint coordinates $\hat{\mathbf{J}}_{\mathrm{3D}} \in \mathbb R^{K \times 3}$ and 3D vertex coordinates $\hat{\mathbf{V}}_{\mathrm{3D}} \in \mathbb R^{N \times 3}$ are predicted.

\noindent \textbf{Coarse-to-Fine Mesh Upsampling.}
Following~\cite{kolotouros2019cmr,lin2021metro,lin2021graphormer}, our FastMETRO estimates a coarse mesh, then upsample the mesh.
In this way, we avoid the redundancy caused by the spatial locality of human mesh vertices.
As in~\cite{kolotouros2019cmr},
FastMETRO obtains the fine mesh output 
$\hat{\mathbf{V}}_{\mathrm{3D}}' \in \mathbb R^{M \times 3}$ 
from the coarse mesh output 
$\hat{\mathbf{V}}_{\mathrm{3D}}$
by performing matrix multiplication with the upsampling matrix $\mathbf{U} \in \mathbb R^{M \times N}$, \ie,
$\hat{\mathbf{V}}_{\mathrm{3D}}' = \mathbf{U}\hat{\mathbf{V}}_{\mathrm{3D}}$,
where the upsampling matrix $\mathbf{U}$ is pre-computed by the sampling algorithm in~\cite{ranjan2018generating}.

\subsection{Training FastMETRO}

\noindent \textbf{3D Vertex Regression Loss.}
To train our model for the regression of 3D mesh vertices, we use $L1$ loss function. 
This regression loss $L^{V}_{\mathrm {3D}}$ is computed by
\begin{align}
    \label{eq:loss_3dvertex}
    L^{V}_{\mathrm {3D}} = \frac{1}{M}
    \Vert \hat{\mathbf{V}}_{\mathrm{3D}}' - \bar{\mathbf{V}}_{\mathrm{3D}} \Vert_{1},
\end{align}
where $\bar{\mathbf{V}}_{\mathrm{3D}} \in \mathbb R^{M \times 3}$ denotes the ground-truth 3D vertex coordinates.

\noindent \textbf{3D Joint Regression Loss.}
In addition to the estimated 3D joints
$\hat{\mathbf{J}}_{\mathrm{3D}}$, we also obtain 3D joints $\hat{\mathbf{J}}_{\mathrm{3D}}' \in \mathbb R^{K \times 3}$ regressed from the fine mesh $\hat{\mathbf{V}}_{\mathrm{3D}}'$, which is the common practice in the literature~\cite{hmrKanazawa17,kolotouros2019cmr,kolotouros2019spin,choi2020pose2mesh,Moon_2020_ECCV_I2L-MeshNet,lin2021metro,lin2021graphormer,Kocabas_PARE_2021,sun2021monocular,dwivedi2021dsr,zhang2021pymaf}.
The regressed joints $\hat{\mathbf{J}}_{\mathrm{3D}}'$ are computed by the matrix multiplication of the joint regression matrix $\mathbf{R} \in \mathbb R^{K \times M}$ and 
the fine mesh
$\hat{\mathbf{V}}_{\mathrm{3D}}'$, \ie, $\hat{\mathbf{J}}_{\mathrm{3D}}' = \mathbf{R} \hat{\mathbf{V}}_{\mathrm{3D}}'$,
where the regression matrix $\mathbf{R}$ is pre-defined in SMPL~\cite{SMPL:2015}.
To train our model for the regression of 3D body joints, we use $L1$ loss function.
This regression loss $L^{J}_{\mathrm {3D}}$ is computed by
\begin{align}
    \label{eq:loss_3djoint}
    L^{J}_{\mathrm {3D}} = \frac{1}{K} (
    \Vert \hat{\mathbf{J}}_{\mathrm{3D}} - \bar{\mathbf{J}}_{\mathrm{3D}} \Vert_{1} + \Vert \hat{\mathbf{J}}_{\mathrm{3D}}' - \bar{\mathbf{J}}_{\mathrm{3D}} \Vert_{1}),
\end{align}
where $\bar{\mathbf{J}}_{\mathrm{3D}} \in \mathbb R^{K \times 3}$ denotes the ground-truth 3D joint coordinates.

\noindent \textbf{2D Joint Projection Loss.}
Following the literature~\cite{hmrKanazawa17,kolotouros2019cmr,kolotouros2019spin,lin2021metro,lin2021graphormer,Kocabas_PARE_2021,sun2021monocular,dwivedi2021dsr,zhang2021pymaf}, for the alignment between the 2D input image and the 3D reconstructed human mesh,
we train our model to estimate weak-perspective camera parameters $\{s, \mathbf{t}\}$;
a scaling factor $s \in \mathbb R$ and a 2D translation vector $\mathbf{t} \in \mathbb R^{2}$.
The weak-perspective camera parameters are estimated from the camera feature obtained by the transformer encoder.
Using the camera parameters, 
we get 2D body joints via an orthographic projection of the estimated 3D body joints.
The projected 2D body joints are computed by
\begin{align}
    \label{eq:loss_2djoint1}
    \hat{\mathbf{J}}_{\mathrm{2D}} = s\mathrm{\Pi}(\hat{\mathbf{J}}_{\mathrm{3D}}) + \mathbf{t},
    \\
    \hat{\mathbf{J}}_{\mathrm{2D}}' = s\mathrm{\Pi}(\hat{\mathbf{J}}_{\mathrm{3D}}') + \mathbf{t},
\end{align}
where $\mathrm{\Pi}(\cdot)$ denotes the orthographic projection;
$\big[\begin{smallmatrix}
  1 & 0 & 0 \\
  0 & 1 & 0
\end{smallmatrix}\big]^{{\scriptscriptstyle\mathsf{T}}} \in \mathbb R^{3 \times 2}$
is used for this projection in FastMETRO.
To train our model with the projection of 3D body joints onto the 2D image, we use $L1$ loss function.
This projection loss $L^{J}_{\mathrm {2D}}$ is computed by
\begin{align}
    \label{eq:loss_2djoin2}
    L^{J}_{\mathrm {2D}} = \frac{1}{K}
    (\Vert \hat{\mathbf{J}}_{\mathrm{2D}} - \bar{\mathbf{J}}_{\mathrm{2D}} \Vert_{1} + \Vert \hat{\mathbf{J}}_{\mathrm{2D}}' - \bar{\mathbf{J}}_{\mathrm{2D}} \Vert_{1}),
\end{align}
where $\bar{\mathbf{J}}_{\mathrm{2D}} \in \mathbb R^{K \times 2}$ denotes the ground-truth 2D joint coordinates.

\noindent \textbf{Total Loss.}
Following the literature~\cite{hmrKanazawa17,kolotouros2019cmr,kolotouros2019spin,kocabas2020vibe,choi2020pose2mesh,Moon_2020_ECCV_I2L-MeshNet,lin2021metro,lin2021graphormer,Kocabas_PARE_2021,sun2021monocular,dwivedi2021dsr,zhang2021pymaf},
we train our model with multiple 3D and 2D training datasets to improve its accuracy and robustness.
This total loss $L_{\mathrm {total}}$ is computed by
\begin{align}
    \label{eq:loss_total}
    L_{\mathrm {total}} = \alpha (\lambda^{V}_{\mathrm {3D}} L^{V}_{\mathrm {3D}} + \lambda^{J}_{\mathrm {3D}} L^{J}_{\mathrm {3D}})
    + \beta \lambda^{J}_{\mathrm {2D}} L^{J}_{\mathrm {2D}},
\end{align}
where $\lambda^{V}_{\mathrm {3D}}, \lambda^{J}_{\mathrm {3D}}, \lambda^{J}_{\mathrm {2D}} > 0$ are loss coefficients and $\alpha, \beta \in \{0,1\}$ are binary flags which denote the availability of ground-truth 3D and 2D coordinates.
\begin{table}[t!]
    \centering
    \caption{
        Configurations for the variants of FastMETRO.
        Each has the same transformer architecture with a different number of layers.
        Only transformer parts are described.
    }
    \resizebox{\textwidth}{!}{
        \begin{tabular}{lccccccc}
        \hline
        \multicolumn{1}{c}{}
            & \multicolumn{1}{c}{}
            & \multicolumn{1}{c}{}
            & \multicolumn{2}{c}{\;\;Enc--1 \& Dec--1\;\;}  
            & \multicolumn{1}{c}{\;\;}
            & \multicolumn{2}{c}{\;\;Enc--2 \& Dec--2\;\;}  
        \\
        \cline{4-5}
        \cline{7-8}
        Model
            & \,\#Params \,
            & Time\,(ms)\;
            & \,\#Layers
            & \,Dimension\,
            & 
            & \,\#Layers
            & \,Dimension\,
        \\
        \hline
        \hline
            FastMETRO--S
            & 9.2M & \; 9.6 \; & 1 & 512 & & 1 & 128
        \\
            FastMETRO--M
            & 17.1M & \; 15.0 \; & 2 & 512 & & 2 & 128
        \\
            FastMETRO--L
            & 24.9M & \; 20.8 \; & 3 & 512 & & 3 & 128
        \\
        \hline
    \end{tabular}}
    \label{table:config}
\end{table}

\section{Implementation Details}
We implement our proposed method with three variants: \mbox{FastMETRO-S}, \mbox{FastMETRO-M}, FastMETRO-L.
They have the same architecture with a different number of layers in the transformer encoder and decoder.
Table~\ref{table:config} shows the configuration for each variant.
Our transformer encoder and decoder are initialized with Xavier Initialization~\cite{xavier2010init}.
Please refer to the supplementary material for complete implementation details.

\section{Experiments}

\subsection{Datasets}
Following the encoder-based transformers~\cite{lin2021metro,lin2021graphormer}, we train our FastMETRO with \textbf{Human3.6M}~\cite{wang2014h36m}, \textbf{UP-3D}~\cite{lassner2017up3d}, \textbf{MuCo-3DHP}~\cite{mehta2018muco}, \textbf{COCO}~\cite{lin2014coco} and \textbf{MPII}~\cite{mykhaylo2014mpii} training datasets, and evaluate the model on P2 protocol in Human3.6M.
Then, we fine-tune our model with \textbf{3DPW}~\cite{marcard20183dpw} training dataset, and evaluate the model on its test dataset.

\begin{table}[t!]
    \centering
    \caption{
        Comparison with transformers for monocular 3D human mesh recovery on Human3.6M~\cite{wang2014h36m}.
        $\dagger$ and $\ast$ indicate training for 60 epochs and 200 epochs, respectively.
    }
    \resizebox{\textwidth}{!}{
        \begin{tabular}{lccccccccc}
        \hline
        \multicolumn{1}{c}{}
            & \multicolumn{2}{c}{\,CNN Backbone\,}
            & \multicolumn{1}{c}{\;}
            & \multicolumn{2}{c}{\,Transformer\,}
            & \multicolumn{1}{c}{\;}
            & \multicolumn{2}{c}{\,Overall\,}
            & \multicolumn{1}{c}{\;}  
        \\
        \cline{2-3}
        \cline{5-6}
        \cline{8-9}
        Model
            & \#Params
            & \,Time\,(ms) 
            & 
            & \#Params
            & \,Time\,(ms)
            & 
            & \#Params
            & \,FPS\,
            & \,PA-MPJPE $\downarrow$
        \\
        \hline
        \hline
            METRO--R50$^\ast$~\cite{lin2021metro}
            & 23.5M 
            & 7.5 
            & 
            & 102.3M 
            & 24.2 
            &
            & 125.8M 
            & 31.5 
            & 40.6
        \\
            METRO--H64$^\dagger$~\cite{lin2021metro}
            & 128.1M 
            & 49.0 
            &
            & 102.3M 
            & 24.2 
            &
            & 230.4M 
            & 13.7 
            & 38.0
        \\
            METRO--H64$^\ast$~\cite{lin2021metro}
            & 128.1M 
            & 49.0 
            &
            & 102.3M 
            & 24.2 
            &
            & 230.4M 
            & 13.7 
            & 36.7
        \\
            MeshGraphormer--H64$^\dagger$~\cite{lin2021graphormer}
            & 128.1M 
            & 49.0 
            &
            & 98.4M 
            & 24.5
            &
            & 226.5M 
            & 13.6 
            & 35.8
        \\
            MeshGraphormer--H64$^\ast$~\cite{lin2021graphormer}
            & 128.1M 
            & 49.0 
            &
            & 98.4M 
            & 24.5 
            &
            & 226.5M 
            & 13.6 
            & 34.5
        \\
        \hhline{-|-|-|-|-|-|-|-|-|-|}
            \cellcolor{gray!7.5}\textbf{FastMETRO--S--R50$^\dagger$}
            & \cellcolor{gray!7.5}23.5M 
            & \cellcolor{gray!7.5}7.5 
            & \cellcolor{gray!7.5}
            & \cellcolor{gray!7.5}\textbf{9.2M} 
            & \cellcolor{gray!7.5}\textbf{9.6} 
            & \cellcolor{gray!7.5}
            & \cellcolor{gray!7.5}\textbf{32.7M} 
            & \cellcolor{gray!7.5}\textbf{58.5} 
            & \cellcolor{gray!7.5}39.4
        \\
            \cellcolor{gray!7.5}\textbf{FastMETRO--M--R50$^\dagger$}
            & \cellcolor{gray!7.5}23.5M 
            & \cellcolor{gray!7.5}7.5 
            & \cellcolor{gray!7.5}
            & \cellcolor{gray!7.5}17.1M 
            & \cellcolor{gray!7.5}15.0 
            & \cellcolor{gray!7.5}
            & \cellcolor{gray!7.5}40.6M 
            & \cellcolor{gray!7.5}44.4 
            & \cellcolor{gray!7.5}38.6
        \\
            \cellcolor{gray!7.5}\textbf{FastMETRO--L--R50$^\dagger$}
            & \cellcolor{gray!7.5}23.5M 
            & \cellcolor{gray!7.5}7.5 
            & \cellcolor{gray!7.5}
            & \cellcolor{gray!7.5}24.9M 
            & \cellcolor{gray!7.5}20.8 
            & \cellcolor{gray!7.5}
            & \cellcolor{gray!7.5}48.4M 
            & \cellcolor{gray!7.5}35.3 
            & \cellcolor{gray!7.5}37.3
        \\
            \cellcolor{gray!7.5}\textbf{FastMETRO--L--H64$^\dagger$}
            & \cellcolor{gray!7.5}128.1M 
            & \cellcolor{gray!7.5}49.0 
            & \cellcolor{gray!7.5}
            & \cellcolor{gray!7.5}24.9M 
            & \cellcolor{gray!7.5}20.8 
            & \cellcolor{gray!7.5}
            & \cellcolor{gray!7.5}153.0M 
            & \cellcolor{gray!7.5}14.3 
            & \cellcolor{gray!7.5}\textbf{33.7}
        \\
        \hline
    \end{tabular}}
    \label{table:compare_trasnformer}
\end{table}

\begin{figure}[t!]
    \centering
    \includegraphics[width=\columnwidth]{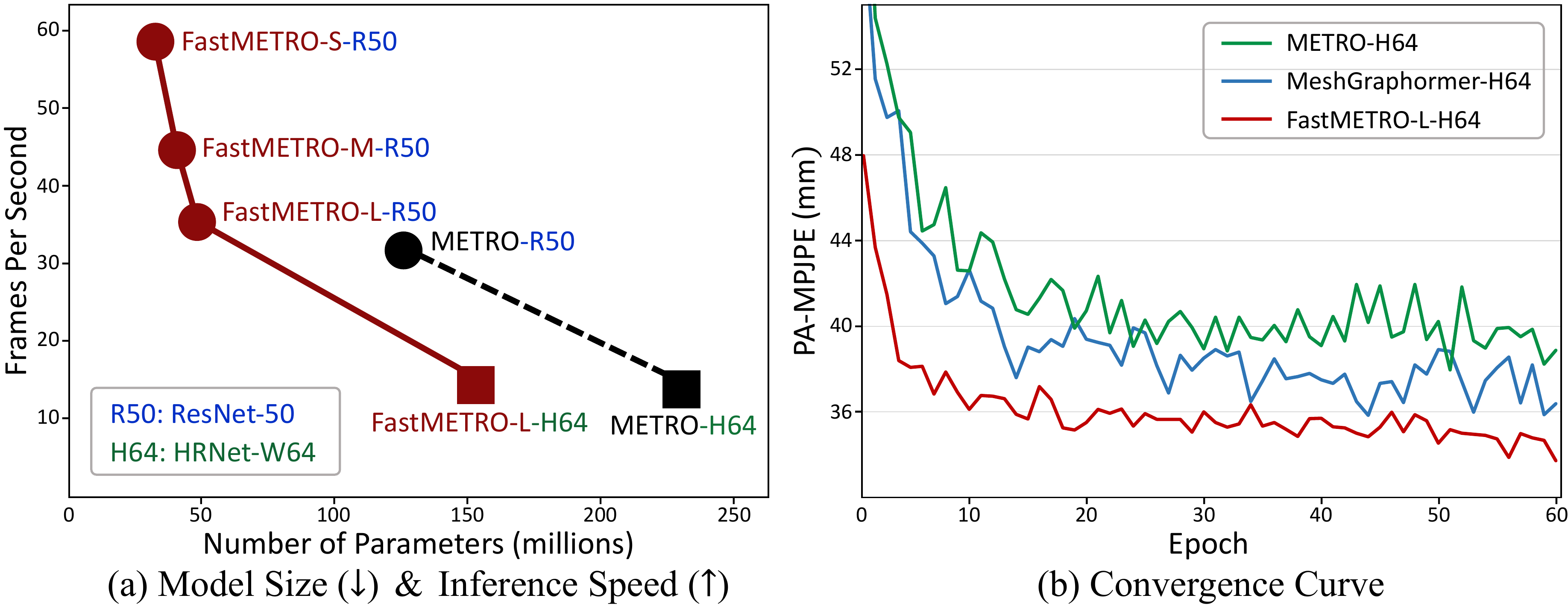}
    \caption{
    Comparison with encoder-based transformers~\cite{lin2021metro,lin2021graphormer} and our proposed models on Human3.6M~\cite{wang2014h36m}.
    The small variant of our FastMETRO shows much faster inference speed, and its large variant converges faster than the transformer encoders.
    }
    \label{fig:convergence}
\end{figure}

Following the common practice~\cite{choi2020pose2mesh,Moon_2020_ECCV_I2L-MeshNet,lin2021metro,lin2021graphormer}, we employ the pseudo 3D human mesh obtained by \mbox{SMPLify-X}~\cite{pavlakos_2019_smplify_x} to train our model with Human3.6M~\cite{wang2014h36m};
there is no available ground-truth 3D human mesh in the Human3.6M training dataset due to the license issue.
For fair comparison, we employ the ground-truth 3D human body joints in Human3.6M during the evaluation of our model.
Regarding the experiments on 3DPW~\cite{marcard20183dpw}, we use its training dataset for fine-tuning our model as in the encoder-based transformers~\cite{lin2021metro,lin2021graphormer}.

\subsection{Evaluation Metrics}
We evaluate our FastMETRO using three evaluation metrics:
MPJPE~\cite{wang2014h36m}, PA-MPJPE~\cite{zhou2018monocap}, MPVPE~\cite{pavlakos_2018_mpve}.
The unit of each metric is millimeter.

\noindent \textbf{MPJPE.}
This metric denotes Mean-Per-Joint-Position-Error.
It measures the Euclidean distances between the predicted and ground-truth joint coordinates.

\noindent \textbf{PA-MPJPE.}
This metric is often called \textit{Reconstruction Error}.
It measures MPJPE after 3D alignment using Procrustes Analysis (PA)~\cite{gower_1975_pa}.

\noindent \textbf{MPVPE.}
This metric denotes Mean-Per-Vertex-Position-Error.
It measures the Euclidean distances between the predicted and ground-truth vertex coordinates.

\begin{table}[t!]
    \centering
    \caption{
        Comparison with the state-of-the-art monocular 3D human pose and mesh recovery methods on 3DPW~\cite{marcard20183dpw} and Human3.6M~\cite{wang2014h36m} among image-based methods.
    }
    \resizebox{\textwidth}{!}{
        \begin{tabular}{lcccccc}
        \hline
        \multicolumn{1}{c}{}
            & \multicolumn{3}{c}{\; 3DPW \;}
            & \multicolumn{1}{c}{\;\;}
            & \multicolumn{2}{c}{\; Human3.6M \;}  
        \\
        \cline{2-4}
        \cline{6-7}
            Model
            & \, MPVPE $\downarrow$ \,
            & \, MPJPE $\downarrow$ \,
            & \, PA-MPJPE $\downarrow$ \,
            & 
            & \, MPJPE $\downarrow$ \,
            & \, PA-MPJPE $\downarrow$ \,
        \\
        \hline
        \hline
            HMR--R50~\cite{hmrKanazawa17}
            & -- 
            & 130.0 
            & 76.7 
            & 
            & 88.0 
            & 56.8
        \\
            GraphCMR--R50~\cite{kolotouros2019cmr}
            & -- 
            & -- 
            & 70.2 
            & 
            & -- 
            & 50.1
        \\
            SPIN--R50~\cite{kolotouros2019spin}
            & 116.4 
            & 96.9 
            & 59.2 
            & 
            & 62.5 
            & 41.1
        \\
            I2LMeshNet--R50~\cite{Moon_2020_ECCV_I2L-MeshNet}
            & -- 
            & 93.2 
            & 57.7 
            & 
            & 55.7 
            & 41.1
        \\
            PyMAF--R50~\cite{zhang2021pymaf}
            & 110.1
            & 92.8
            & 58.9
            & 
            & 57.7 
            & 40.5
        \\
            ROMP--R50~\cite{sun2021monocular}
            & 105.6
            & 89.3
            & 53.5
            & 
            & -- 
            & --
        \\
            ROMP--H32~\cite{sun2021monocular}
            & 103.1
            & 85.5
            & 53.3
            & 
            & -- 
            & --
        \\
            PARE--R50~\cite{Kocabas_PARE_2021}
            & 99.7
            & 82.9 
            & 52.3 
            & 
            & --
            & --
        \\
            METRO--R50~\cite{lin2021metro}
            & --
            & -- 
            & -- 
            & 
            & 56.5 
            & 40.6
        \\
            DSR--R50~\cite{dwivedi2021dsr}
            & 99.5
            & 85.7 
            & 51.7 
            & 
            & 60.9
            & 40.3
        \\
            METRO--H64~\cite{lin2021metro}
            & 88.2 
            & 77.1 
            & 47.9 
            & 
            & 54.0 
            & 36.7
        \\
            PARE--H32~\cite{Kocabas_PARE_2021}
            & 88.6
            & 74.5 
            & 46.5 
            & 
            & --
            & --
        \\
            MeshGraphormer--H64~\cite{lin2021graphormer}
            & 87.7 
            & 74.7 
            & 45.6 
            & 
            & \textbf{51.2} 
            & 34.5
        \\
        \hhline{-|-|-|-|-|-|-|}
            \cellcolor{gray!7.5}\textbf{FastMETRO--S--R50}
            & \cellcolor{gray!7.5}91.9 
            & \cellcolor{gray!7.5}79.6 
            & \cellcolor{gray!7.5}49.3 
            & \cellcolor{gray!7.5}
            & \cellcolor{gray!7.5}55.7 
            & \cellcolor{gray!7.5}39.4
        \\
            \cellcolor{gray!7.5}\textbf{FastMETRO--M--R50}
            & \cellcolor{gray!7.5}91.2 
            & \cellcolor{gray!7.5}78.5 
            & \cellcolor{gray!7.5}48.4 
            & \cellcolor{gray!7.5}
            & \cellcolor{gray!7.5}55.1 
            & \cellcolor{gray!7.5}38.6
        \\
            \cellcolor{gray!7.5}\textbf{FastMETRO--L--R50}
            & \cellcolor{gray!7.5}90.6 
            & \cellcolor{gray!7.5}77.9 
            & \cellcolor{gray!7.5}48.3 
            & \cellcolor{gray!7.5}
            & \cellcolor{gray!7.5}53.9 
            & \cellcolor{gray!7.5}37.3
        \\
            \cellcolor{gray!7.5}\textbf{FastMETRO--L--H64}
            & \cellcolor{gray!7.5}\textbf{84.1} 
            & \cellcolor{gray!7.5}\textbf{73.5} 
            & \cellcolor{gray!7.5}\textbf{44.6} 
            & \cellcolor{gray!7.5}
            & \cellcolor{gray!7.5}52.2 
            & \cellcolor{gray!7.5}\textbf{33.7}
        \\
        \hline
    \end{tabular}}
    \label{table:previous}
\end{table}

\subsection{Experimental Results}
We evaluate the model-size variants of our FastMETRO on the 3DPW~\cite{marcard20183dpw} and Human3.6M~\cite{wang2014h36m} datasets.
In this paper, the inference time is measured using a single NVIDIA V100 GPU with a batch size of 1.

\noindent \textbf{Comparison with Encoder-Based Transformers.}
In Table~\ref{table:compare_trasnformer}, we compare our models with METRO~\cite{lin2021metro} and Mesh Graphormer~\cite{lin2021graphormer} on the Human3.6M~\cite{wang2014h36m} dataset.
Note that encoder-based transformers~\cite{lin2021metro,lin2021graphormer} are implemented with \mbox{ResNet-50}~\cite{resnet} (\textbf{R50}) or
\mbox{HRNet-W64}~\cite{wang2019hrnet} (\textbf{H64}).
\mbox{FastMETRO-S} outperforms METRO when both models employ the same CNN backbone (R50), 
although our model demands only 8.99\% of the parameters in the transformer architecture.
Regarding the \mbox{overall} inference speed, our model is 1.86$\times$ faster.
It is worth noting that 
\mbox{FastMETRO-L-R50} achieves similar results with \mbox{METRO-H64}, but our model is 2.58$\times$ faster. 
\mbox{FastMETRO-L} outperforms \mbox{Mesh Graphormer} 
when both models employ the same CNN backbone (H64), 
while our model demands only 25.30\% of the parameters in the transformer architecture.
Also, our model converges much faster than the encoder-based methods as shown in Figure~\ref{fig:convergence}.

\noindent \textbf{Comparison with Image-Based Methods.}
In Table~\ref{table:previous}, we compare our \mbox{FastMETRO} with the image-based methods for 3D human mesh reconstruction on 3DPW~\cite{marcard20183dpw} and Human3.6M~\cite{wang2014h36m}.
Note that existing methods are implemented with
\mbox{ResNet-50}~\cite{resnet} (\textbf{R50}) or
\mbox{HRNet-W32}~\cite{wang2019hrnet} (\textbf{H32}) or 
\mbox{HRNet-W64}~\cite{wang2019hrnet} (\textbf{H64}).
When all models employ R50 as their CNN backbones,
\mbox{FastMETRO-S} achieves the best results without 
iterative fitting procedures or test-time optimizations.
\mbox{FastMETRO-L-H64} achieves the state of the art in every evaluation metric on the 3DPW dataset and PA-MPJPE metric on the Human3.6M dataset.

\begin{figure}[t!]
    \centering
    \includegraphics[width=\columnwidth]{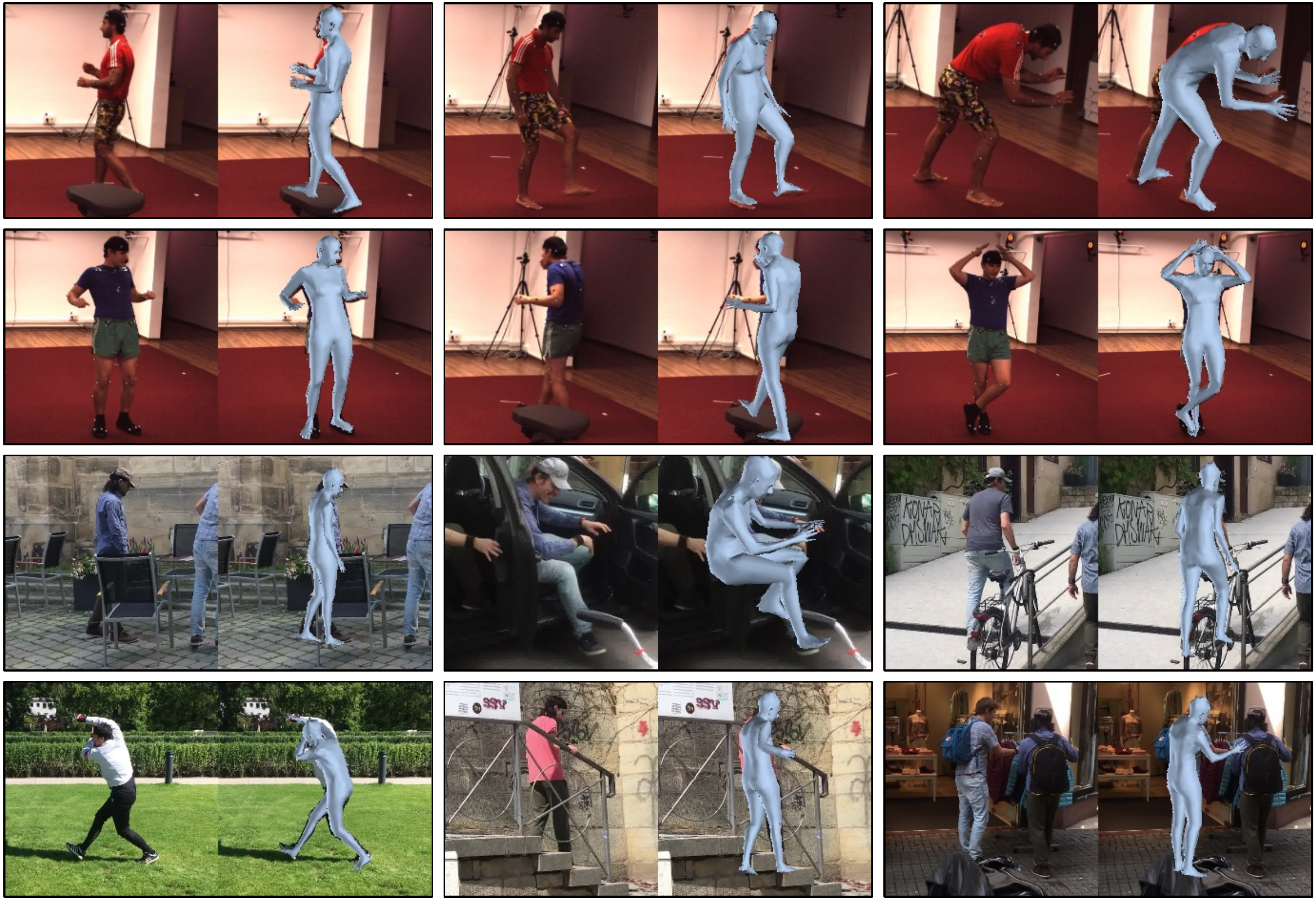}
    \caption{
    Qualitative results of our FastMETRO on Human3.6M~\cite{wang2014h36m} and 3DPW~\cite{marcard20183dpw}. 
    We visualize the 3D human mesh estimated by FastMETRO-L-H64.
    By leveraging the attention mechanism in the transformer,
    our model is robust to partial occlusions.
    }
    \label{fig:qualitative_human}
\end{figure}

\begin{figure}[t!]
    \centering
    \includegraphics[width=\columnwidth]{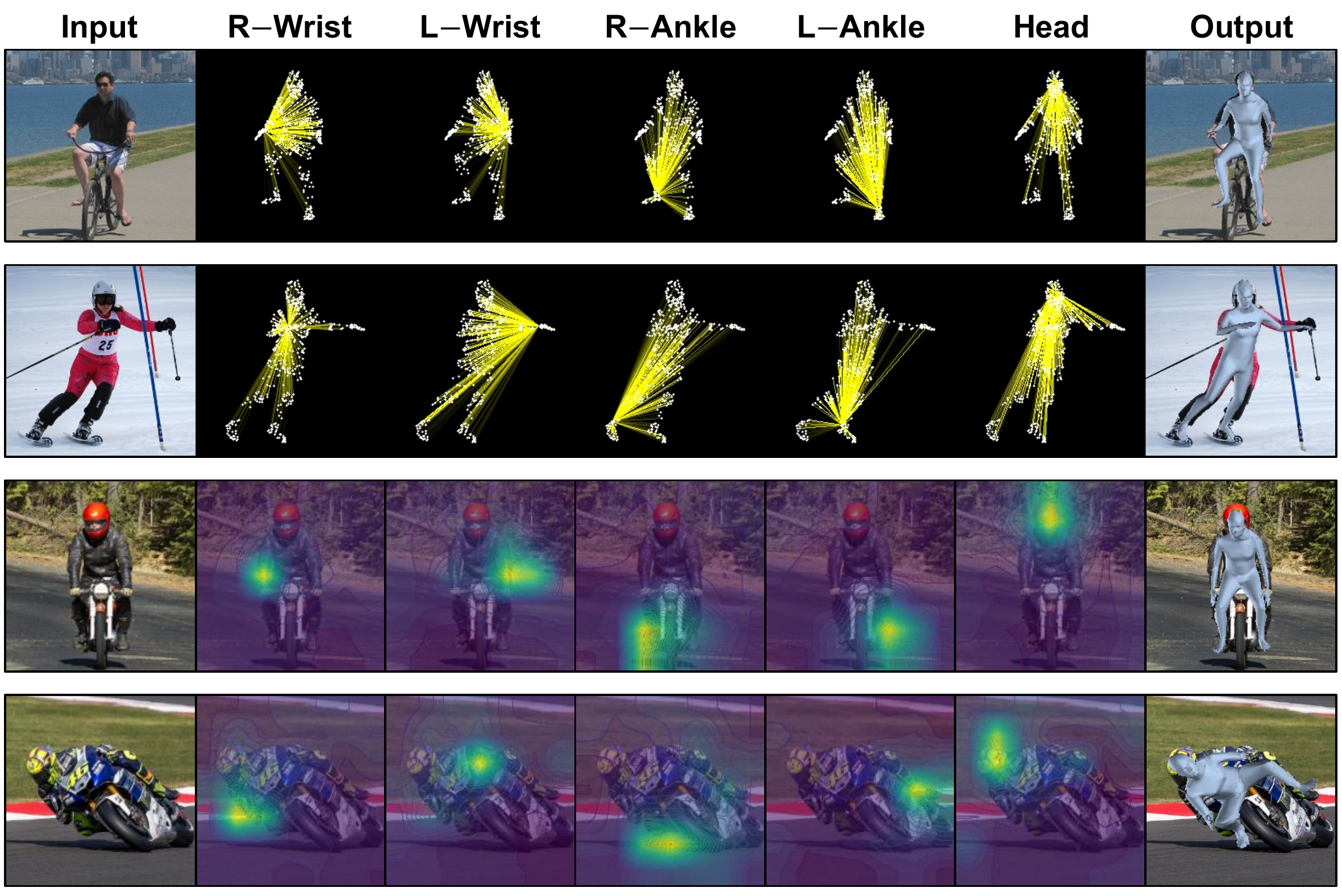}
    \caption{
    Qualitative results of FastMETRO-L-H64 on COCO~\cite{lin2014coco}.
    We visualize the attentions scores in \mbox{self-attentions} (top two rows) and \mbox{cross-attentions} (bottom two rows).
    The brighter lines or regions indicate higher attention scores.
    }
    \label{fig:qualitative_self_cross_attn}
\end{figure}

\noindent \textbf{Visualization of Self-Attentions.}
In Figure~\ref{fig:qualitative_self_cross_attn}, the first and second rows show the visualization of the attention scores in \mbox{self-attentions} between a specified body joint and mesh vertices.
We obtain the scores by averaging attention scores from all attention heads of all \mbox{multi-head} \mbox{self-attention} modules in our transformer decoder.
As shown in Figure~\ref{fig:qualitative_self_cross_attn}, our FastMETRO effectively captures the \mbox{non-local} relations among joints and vertices via \mbox{self-attentions} in the transformer.
This improves the robustness to environment variations such as occlusions.

\noindent \textbf{Visualization of Cross-Attentions.}
In Figure~\ref{fig:qualitative_self_cross_attn}, the third and fourth rows show the visualization of the attention scores in \mbox{cross-attentions} between a specified body joint and image regions.
We obtain the scores by averaging attention scores from all attention heads of all \mbox{multi-head} \mbox{cross-attention} modules in our transformer decoder.
As shown in Figure~\ref{fig:qualitative_self_cross_attn}, the input tokens used in our transformer decoder focus on their relevant image regions.
By leveraging the \mbox{cross-attentions} between disentangled modalities, 
our \mbox{FastMETRO} effectively learns to regress the 3D coordinates of joints and vertices from a 2D image.

\subsection{Ablation Study}
We analyze the effects of different components in our FastMETRO as shown in Table~\ref{table:ablation}.
Please refer to the supplementary material for more experiments.

\noindent \textbf{Attention Masking.}
To effectively learn the local relations among mesh vertices, \mbox{GraphCMR~\cite{kolotouros2019cmr}} and Mesh Graphormer~\cite{lin2021graphormer} perform graph convolutions based on the topology of SMPL human triangle mesh~\cite{SMPL:2015}.
For the same goal,
we mask self-attentions of non-adjacent vertices according to the topology.
When we evaluate our model without masking the attentions,
the regression accuracy drops as shown in the first row of Table~\ref{table:ablation}.
This demonstrates that masking the attentions of non-adjacent vertices is effective.
To compare the effects of attention masking with graph convolutions,
we train our model using graph convolutions without masking the attentions.
As shown in the second row of Table~\ref{table:ablation}, we obtain similar results but this requires more parameters.
We also evaluate our model when we mask the attentions in half attention heads, \ie, there is no attention masking in other half attention heads.
In this case,
we get similar results using the same number of parameters as shown in the third row of Table~\ref{table:ablation}.

\begin{table}[t!]
    \centering
    \caption{
        Ablation study of our FastMETRO on Human3.6M~\cite{wang2014h36m}. 
        The effects of different components are evaluated.
        The default model is FastMETRO-S-R50.
    }
    \resizebox{\textwidth}{!}{
        \begin{tabular}{lccc}
        \hline
        \multicolumn{1}{c}{}
            & \multicolumn{1}{c}{\;\;}
            & \multicolumn{2}{c}{\; Human3.6M \;}  
        \\
        \cline{3-4}
            Model
            & \; \#Params \;
            & \, MPJPE $\downarrow$
            & \, PA-MPJPE $\downarrow$ \,
        \\
        \hline
        \hline
            w/o attention masking
            & \textbf{32.7M} 
            & 58.0
            & 40.7
        \\
            w/o attention masking + w/ graph convolutions
            & 33.1M
            & 56.6
            & \textbf{39.4}
        \\
            w/ attention masking in half attention heads
            & \textbf{32.7M}
            & 55.8
            & \textbf{39.4}
        \\
            w/ learnable upsampling layers
            & 45.4M 
            & 58.1
            & 41.1
        \\
            w/o progressive dimensionality reduction
            & 39.5M 
            & \textbf{55.5}
            & 39.6
        \\
        \hhline{-|-|-|-|}
            \cellcolor{gray!7.5}\textbf{FastMETRO--S--R50}
            & \cellcolor{gray!7.5}\textbf{32.7M} 
            & \cellcolor{gray!7.5}55.7
            & \cellcolor{gray!7.5}\textbf{39.4}
        \\
        \hline
    \end{tabular}}
    \label{table:ablation}
\end{table}

\noindent \textbf{Coarse-to-Fine Mesh Upsampling.}
The existing transformers~\cite{lin2021metro,lin2021graphormer} also first estimate a coarse mesh, then upsample the mesh to obtain a fine mesh.
They employ two learnable linear layers for the upsampling.
In our FastMETRO, we use the pre-computed upsampling matrix $\mathbf U$ to reduce optimization difficulty as in~\cite{kolotouros2019cmr};
this upsampling matrix is a sparse matrix which has only about $25K$ \mbox{non-zero} elements.
When we perform the mesh upsampling using learnable linear layers instead of the matrix $\mathbf U$,
the regression accuracy drops as shown in the fourth row of Table~\ref{table:ablation},
although it demands much more parameters.

\noindent \textbf{Progressive Dimensionality Reduction.}
Following the existing transformer encoders~\cite{lin2021metro,lin2021graphormer}, 
we also progressively reduce the hidden dimension sizes in our transformer via linear projections.
To evaluate its effectiveness, 
we train our model using the same number of transformer layers but without progressive dimensionality reduction, \ie, hidden dimension sizes in all transformer layers are the same.
As shown in the fifth row of Table~\ref{table:ablation},
we obtain similar results but this requires much more parameters.
This demonstrates that the dimensionality reduction is helpful for our model to achieve decent results using fewer parameters.

\noindent \textbf{Generalizability.}
Our model can reconstruct any arbitrary 
3D objects by changing the number of input tokens used in the transformer 
decoder.
Note that we can employ learnable layers for coarse-to-fine mesh upsampling without masking attentions.
For 3D hand mesh reconstruction, there is a pre-computed upsampling matrix and a human hand model such as MANO~\cite{MANO:SIGGRAPHASIA:2017}. 
Thus, we can leverage the 
\begin{wrapfigure}{r}{0.5\textwidth}
    \centering
    \includegraphics[width=.9\linewidth]{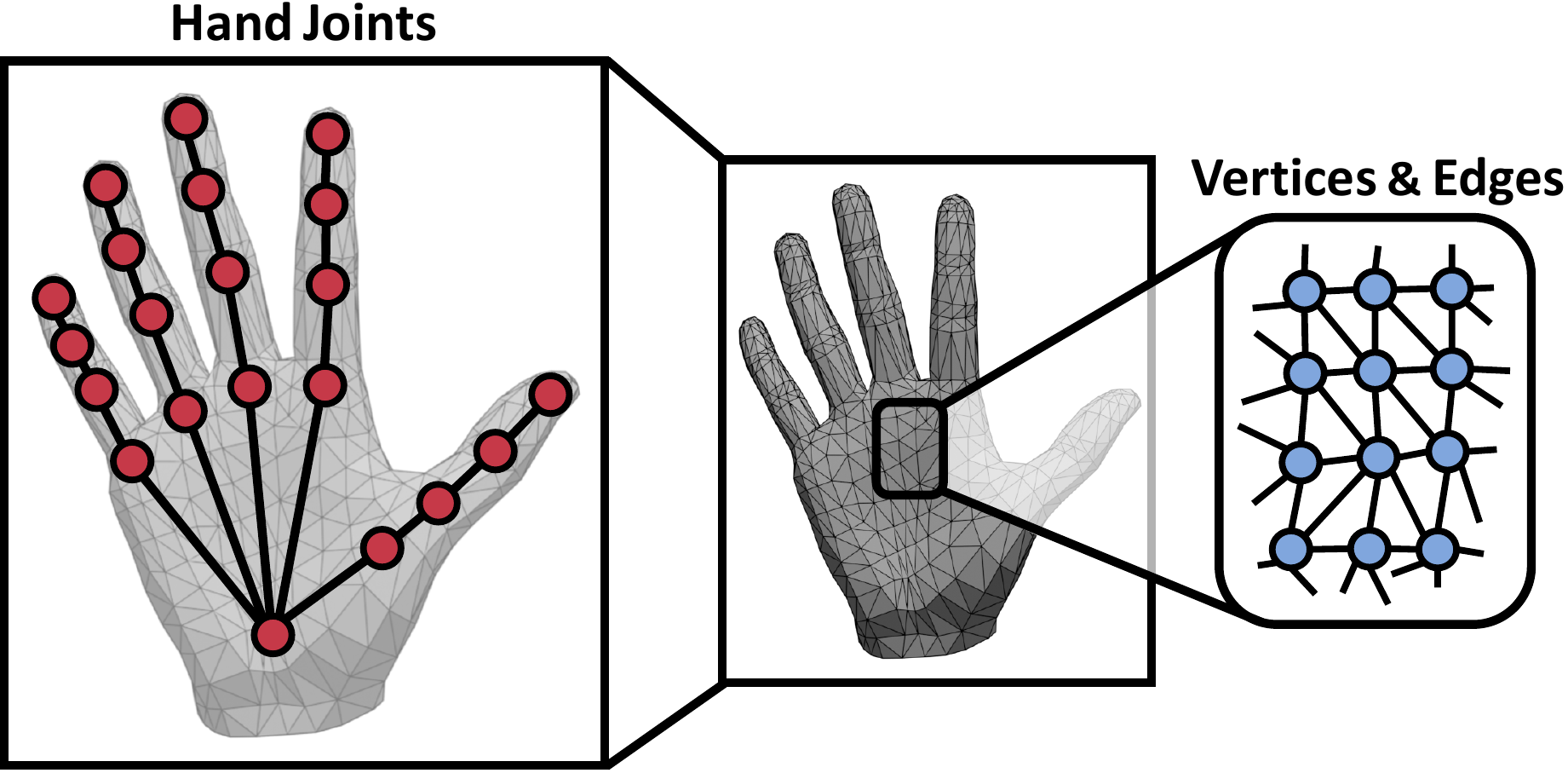}
    \caption{
        Hand Joints and Mesh Topology.
    }
    \label{fig:hand_mesh_topology}
\end{wrapfigure}
matrix for mesh upsampling and mask self-attentions of non-adjacent vertices in the same way with 3D human mesh recovery.
As illustrated in Figure~\ref{fig:hand_mesh_topology}, we can obtain an adjacency matrix and construct 
joint and vertex tokens from the human hand mesh topology. 
To validate the generalizability of our method,
we train FastMETRO-L-H64 on the \mbox{FreiHAND}~\cite{zimmermann2019freihand} training
dataset and evaluate the model.
As shown in Table~\ref{table:freihand}, our proposed model achieves competitive results on FreiHAND.

\begin{table}[t!]
    \centering
    \caption{
        Comparison with transformers for monocular 3D hand mesh recovery on FreiHAND~\cite{zimmermann2019freihand}.
        Test-time augmentation is not applied to these transformers.
    }
    \resizebox{\textwidth}{!}{
        \begin{tabular}{lccccccc}
        \hline
        \multicolumn{1}{c}{}
        & \multicolumn{1}{c}{\;\;}
        & \multicolumn{1}{c}{\, Transformer \,}
        & \multicolumn{1}{c}{\;\;}
        & \multicolumn{1}{c}{\, Overall \,}
        & \multicolumn{1}{c}{\;\;}
        & \multicolumn{2}{c}{\; FreiHAND \;}  
        \\
        \cline{3-3}
        \cline{5-5}
        \cline{7-8}
            Model
            & 
            & \, \#Params \,
            &
            & \, \#Params \,
            &
            & \, PA-MPJPE $\downarrow$ \,
            & \, F@15mm $\uparrow$ \,
        \\
        \hline
        \hline
            METRO--H64~\cite{lin2021metro}
            & 
            & 102.3M
            &
            & 230.4M
            &
            & 6.8
            & 0.981
        \\
        \hhline{-|-|-|-|-|-|-|-|}
            \cellcolor{gray!7.5}\textbf{FastMETRO--L--H64}
            & \cellcolor{gray!7.5} 
            & \cellcolor{gray!7.5}\textbf{24.9M}
            & \cellcolor{gray!7.5}
            & \cellcolor{gray!7.5}\textbf{153.0M}
            & \cellcolor{gray!7.5}
            & \cellcolor{gray!7.5}\textbf{6.5}
            & \cellcolor{gray!7.5}\textbf{0.982}
        \\
        \hline
    \end{tabular}}
    \label{table:freihand}
\end{table}

\begin{figure}[t!]
    \centering
    \includegraphics[width=\columnwidth]{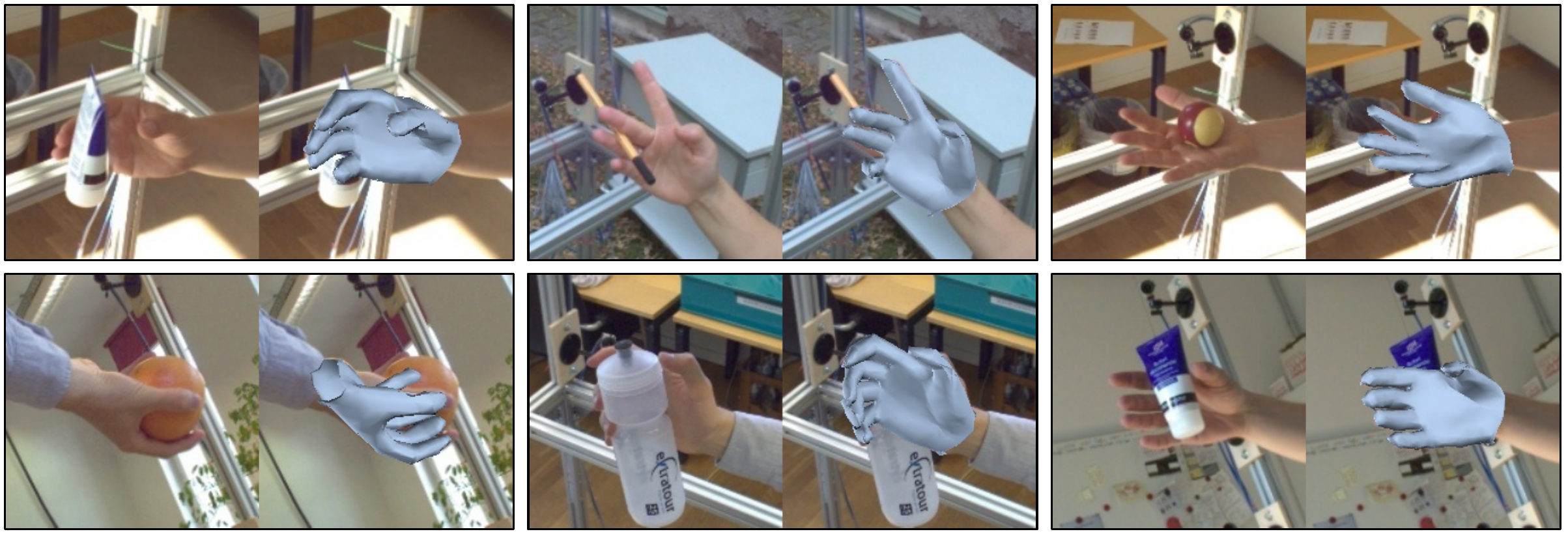}
    \caption{
    Qualitative results of our FastMETRO on FreiHAND~\cite{zimmermann2019freihand}. 
    We visualize the 3D hand mesh estimated by FastMETRO-L-H64.
    By leveraging the attention mechanism in the transformer,
    our model is robust to partial occlusions.
    }
    \label{fig:qualitative_hand}
\end{figure}
\section{Conclusion}
We identify the performance bottleneck in the encoder-based transformers is due to the design of input tokens, and resolve this issue via an \mbox{encoder-decoder} architecture.
This allows our model to demand much fewer parameters and shorter inference time, which is more appropriate for practical use.
The proposed method leverages the human body prior encoded in SMPL human mesh, which reduces optimization difficulty and leads to faster convergence with higher accuracy. 
To be specific, we mask self-attentions of non-adjacent vertices and perform coarse-to-fine mesh upsampling.
We demonstrate that our method improves the Pareto-front of accuracy and efficiency.
Our FastMETRO achieves the robustness to occlusions by capturing \mbox{non-local} relations among body joints and mesh vertices, which outperforms \mbox{image-based} methods on the Human3.6M and 3DPW datasets.
A limitation is that a substantial number of samples are required to train our model as in the encoder-based transformers.

{

{\noindent \textbf{Acknowledgments.}
This work was supported by Institute of Information \& communications Technology Planning \& Evaluation (IITP)
grant funded by the Korea government (MSIT) 
(No. 2022-0-00290, Visual Intelligence for Space-Time Understanding and Generation based on Multi-layered Visual Common Sense;
and No. 2019-0-01906, Artificial Intelligence Graduate School Program (POSTECH)).
}}

\bibliographystyle{splncs04}
\bibliography{arxiv_main}

\clearpage
\setcounter{section}{0}
\setcounter{figure}{0}
\setcounter{table}{0}
\renewcommand{\thesection}{\Alph{section}}
\renewcommand\thesubsection{\thesection.\arabic{subsection}}

\title{
Cross-Attention of Disentangled Modalities for 3D Human Mesh Recovery with Transformers\\
\vspace{2mm}
\textmd{--- Supplementary Material ---}} 
\titlerunning{FastMETRO} 
\authorrunning{J. Cho et al.} 
\author{}
\institute{}

\maketitle
In this supplementary material, 
we present more implementation details \mbox{(Section~\ref{supp:implementation})},
quantitative evaluations \mbox{(Section~\ref{supp:quntitative})}
and qualitative evaluations \mbox{(Section~\ref{supp:qualitative})}, 
which are not included in the main paper due to its limited space.

\vspace{7mm}
\section{Implementation Details}
\label{supp:implementation}
\vspace{2.5mm}
PyTorch~\cite{pytorch} is used to implement our FastMETRO.
We employ ResNet-50~\cite{resnet} or HRNet-W64~\cite{wang2019hrnet} as our CNN backbone,
where each backbone is initialized with ImageNet~\cite{deng2009imagenet} pre-trained weights.
For the initialization of our transformer, we use Xavier Initialization~\cite{xavier2010init}.
Given an input image of size $224 \times 224 \times 3$, our CNN backbone produces the image features $\mathbf X_{I} \in \mathbb R^{H \times W \times C}$, where $H=W=7$ and $C=2048$.
The hidden dimension size of the camera token is $D=512$, which is also same for each joint token $\mathbf{t}^{J}_{i}$ and vertex token $\mathbf{t}^{V}_{j}$.
Following~\cite{lin2021metro,lin2021graphormer}, the number of joint tokens is $K=14$ and that of vertex tokens is $N=431$.
The number of vertices in the fine mesh output $\hat{\mathbf{V}}_{\mathrm{3D}}'$ is $M=6890$, which is same with that of SMPL~\cite{SMPL:2015}.
The number of heads in multi-head attention modules is $8$.
We employ two linear layers with a ReLU activation function for the MLPs in our transformer layers.
Regarding the 3D coordinates regressor or camera predictor, we use a linear layer.
To retain spatial information for the flattened image features $\mathbf X_{F}$, we use fixed sine positional encodings as in~\cite{carion2020end}.
Note that the \mbox{pre-computed} matrix for mesh upsampling and the adjacency matrix for attention masking are sparse matrices in our implementation;
only about $25K$ elements are \mbox{non-zeros} in the upsampling matrix and about $3K$ elements are \mbox{non-zeros} in the adjacency matrix.
We leverage these sparse matrices for memory-efficient implementation.

We use AdamW optimizer~\cite{loshchilov2018decoupled} with the learning rate of $10^{-4}$, weight decay of $10^{-4}$, $\beta_1=0.9$ and $\beta_2=0.999$.
For stable training of our transformer, we apply gradient clipping and set the maximal gradient norm value to $0.3$.
The loss coefficients are
$\lambda^{V}_{\mathrm {3D}} = \lambda^{J}_{\mathrm {2D}} = 100$ and $\lambda^{J}_{\mathrm {3D}} = 1000$.
We train our FastMETRO with a batch size of $16$ for $60$ epochs, which takes about 4 days on 4 NVIDIA V100 GPUs (16GB RAM).
Note that METRO~\cite{lin2021metro} and Mesh Graphormer~\cite{lin2021graphormer} are trained with a batch size of 32 for 200 epochs, which takes about 5 days on 8 NVIDIA V100 GPUs (32GB RAM). 
During training, we apply the standard data augmentation for this task as in~\cite{kolotouros2019spin,Moon_2020_ECCV_I2L-MeshNet,lin2021metro,lin2021graphormer}.
\newpage
\section{Quantitative Evaluations}
\renewcommand\thefigure{B\arabic{figure}}
\renewcommand\thetable{B\arabic{table}}
\label{supp:quntitative}
\begin{table}[t!]
    \centering
    \caption{
        Ablation study of our FastMETRO on Human3.6M~\cite{wang2014h36m}. 
        The effects of different components are evaluated.
        The default model is FastMETRO-S-R50.
    }
    \resizebox{\textwidth}{!}{
        \begin{tabular}{lccc}
        \hline
        \multicolumn{1}{c}{}
            & \multicolumn{1}{c}{\;\;}
            & \multicolumn{2}{c}{\; Human3.6M \;}  
        \\
        \cline{3-4}
            Model
            & \; \#Params \;
            & \, MPJPE $\downarrow$
            & \, PA-MPJPE $\downarrow$ \,
        \\
        \hline
        \hline
            w/ baseline setting
            & 51.9M
            & 59.0
            & 41.8
        \\
            w/ learnable positional encodings
            & 32.8M
            & 56.2
            & 39.7
        \\
            w/ camera token in transformer decoder
            & \textbf{32.7M}
            & 56.1
            & 39.5
        \\
            w/ camera prediction using estimated mesh
            & 32.8M
            & 58.0
            & 39.6
        \\
        \hhline{-|-|-|-|}
            \cellcolor{gray!7.5}\textbf{FastMETRO--S--R50}
            & \cellcolor{gray!7.5}\textbf{32.7M} 
            & \cellcolor{gray!7.5}\textbf{55.7}
            & \cellcolor{gray!7.5}\textbf{39.4}
        \\
        \hline
    \end{tabular}}
    \label{table:supp_ablation}
\end{table}

\noindent \textbf{Baseline.}
To validate the effectiveness of our method, 
we evaluate a na\"ive transformer \mbox{encoder-decoder} architecture.
Following encoder-based methods~\cite{lin2021metro,lin2021graphormer},
we employ a 3D human mesh for input tokens and learnable layers for mesh upsampling, and do not perform attention masking.
For the construction of joint and vertex tokens, we linearly project the 3D coordinates of joints and vertices in the human mesh.
To simplify this model, we do not progressively reduce hidden dimension sizes in the transformer.
This baseline shows lower regression accuracy as shown in the first row of Table~\ref{table:supp_ablation}, 
although it demands much more parameters.
This demonstrates that our FastMETRO effectively improves the accuracy and reduces the number of parameters required in the architecture. 

\noindent \textbf{Positional Encodings.}
Following~\cite{carion2020end}, we employ fixed sine positional encodings for retaining spatial information in our transformer.
When we use learnable embeddings as positional encodings,
we obtain similar results but this demands more parameters as shown in the second row of Table~\ref{table:supp_ablation}.

\noindent \textbf{Weak-Perspective Camera Parameters.}
We estimate these parameters using a camera token in our transformer encoder.
On the other hand,
GraphCMR~\cite{kolotouros2019cmr} estimates the camera parameters from the vertex features obtained by graph convolutional layers, and encoder-based transformers~\cite{lin2021metro,lin2021graphormer} predict the camera parameters from the 3D coordinates of mesh vertices estimated by transformer layers.
As shown in the third and fourth rows of Table~\ref{table:supp_ablation}, 
we also evaluate our \mbox{FastMETRO} using different methods to predict the camera parameters.
When we employ a camera token in our transformer decoder,
we obtain similar results.
When we predict the camera parameters using the 3D mesh estimated by transformer layers, the regression accuracy drops.

\renewcommand\thefigure{C\arabic{figure}}
\renewcommand\thetable{C\arabic{table}}

\begin{figure}[t!]
    \centering
    \includegraphics[width=\textwidth]{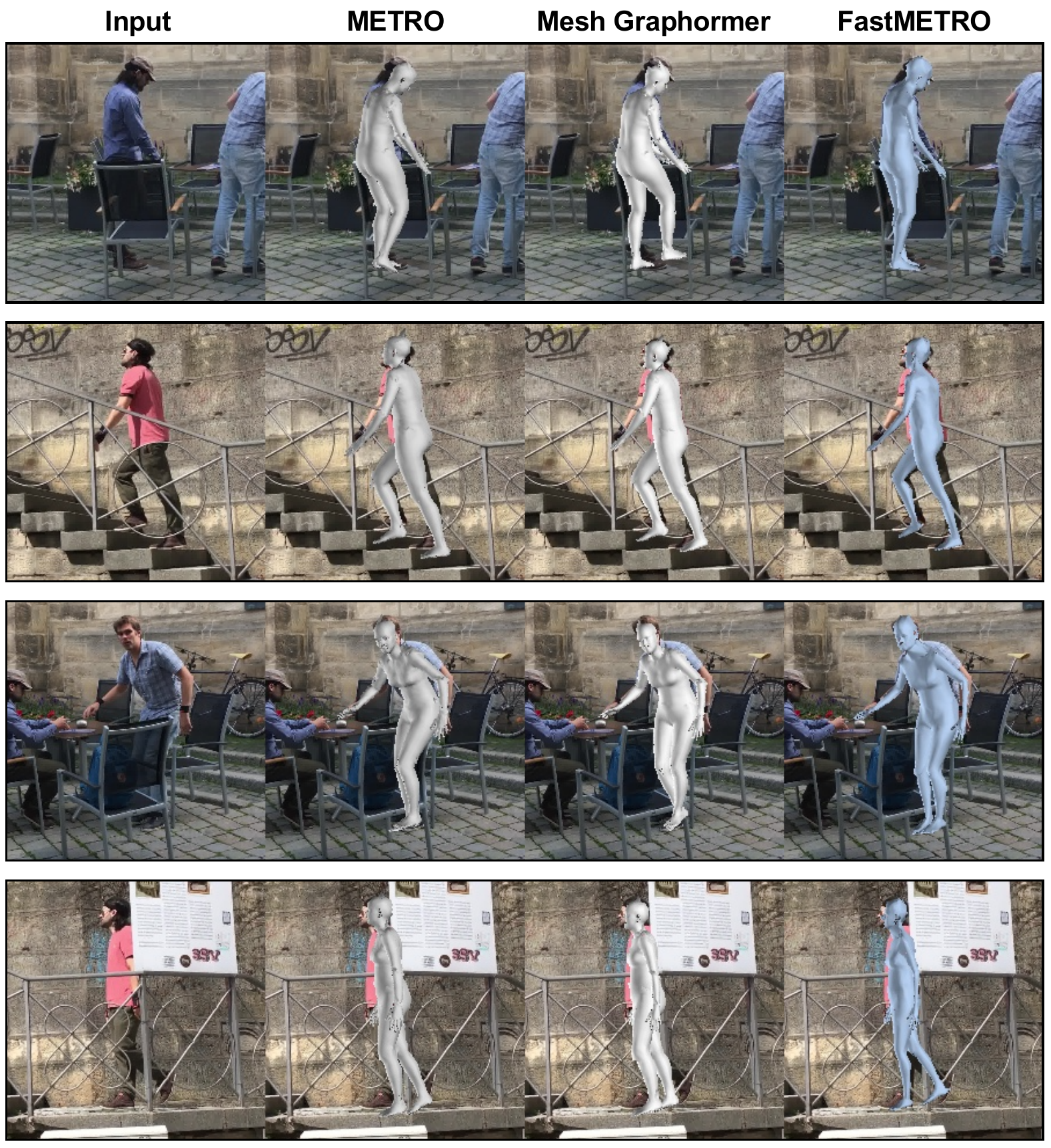}
    \vspace{-4mm}
    \caption{
    Comparison with encoder-based transformers~\cite{lin2021metro,lin2021graphormer} and FastMETRO-L-H64 on 3DPW~\cite{marcard20183dpw}.
    Our model achieves competitive results using much fewer parameters, and shows more favorable body pose especially for knees and ankles.
    }
    \vspace{-2mm}
    \label{fig:supp_qualitative_transformers}
\end{figure}

\begin{figure}[t!]
    \centering
    \includegraphics[width=\textwidth]{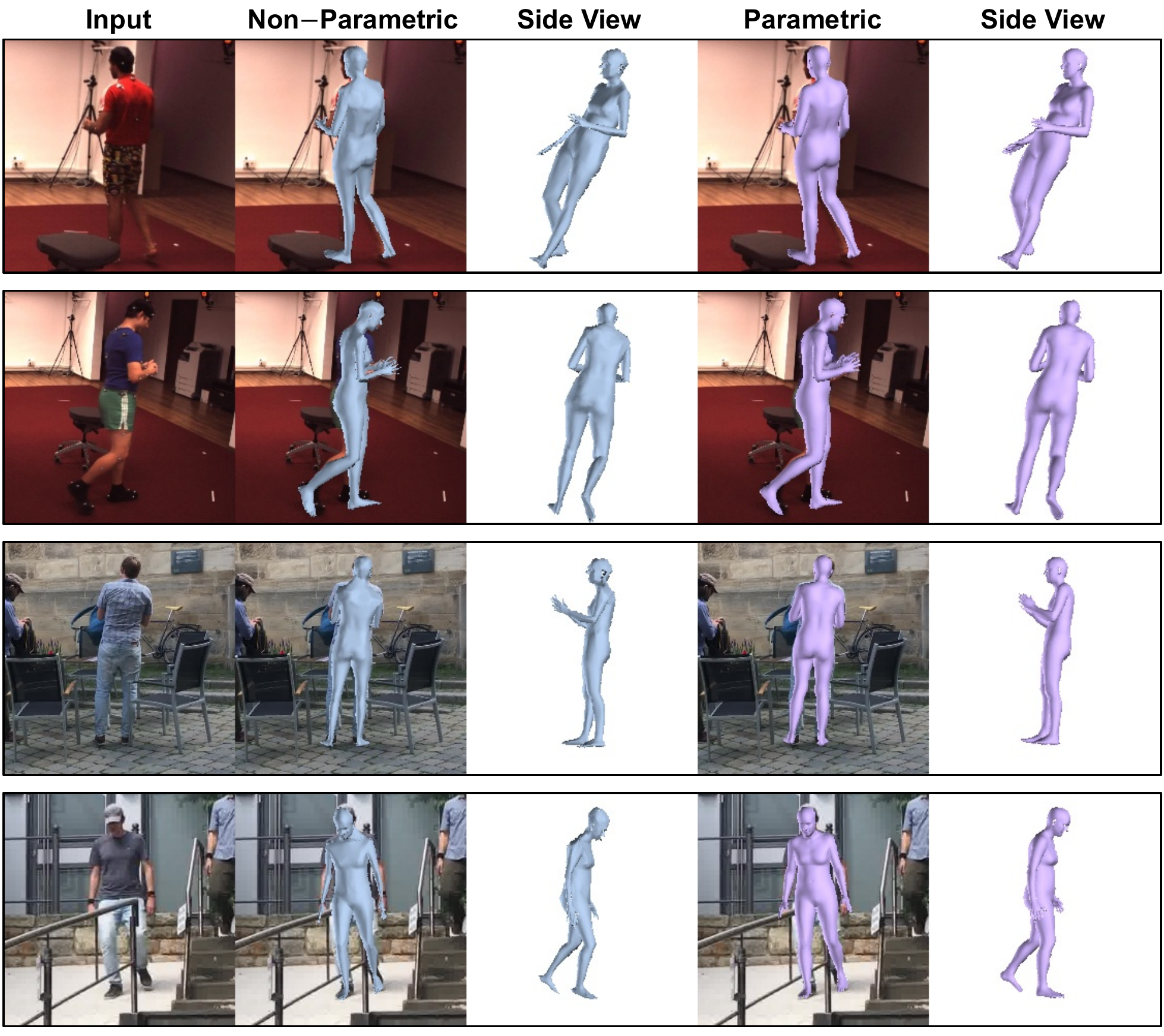}
    \vspace{-4mm}
    \caption{
    Qualitative results of FastMETRO-L-H64 on Human3.6M~\cite{wang2014h36m} and 3DPW~\cite{marcard20183dpw}.
    We can optionally learn to regress SMPL~\cite{SMPL:2015} parameters from the 3D coordinates of mesh vertices estimated by our FastMETRO.
    }
    \label{fig:supp_qualitative_smpl}
\end{figure}

\section{Qualitative Evaluations}
\label{supp:qualitative}

\noindent \textbf{Comparison with Encoder-Based Transformers.}
Figure~\ref{fig:supp_qualitative_transformers} shows the qualitative comparison of transformer encoders~\cite{lin2021metro,lin2021graphormer} with our method. 
As shown in Figure~\ref{fig:supp_qualitative_transformers},
FastMETRO-L-H64 achieves competitive results,
although our model requires only about 25\% of the parameters in the transformer architecture compared with the encoder-based transformers.
Note that FastMETRO captures more detailed body pose especially for knees and ankles.

\noindent \textbf{SMPL Parameters from Estimated Mesh.}
As in~\cite{kolotouros2019cmr,Moon_2020_ECCV_I2L-MeshNet},
we can optionally regress SMPL~\cite{SMPL:2015} parameters from the output mesh estimated by our model. 
To be specific, we first regress the 3D coordinates of human mesh vertices via our model, then predict SMPL pose and shape coefficients via a SMPL parameter regressor which takes the estimated 3D mesh vertices as input.
Following~\cite{kolotouros2019cmr,Moon_2020_ECCV_I2L-MeshNet}, we employ fully connected layers with skip connections as the SMPL parameter regressor.
In this way, we can reconstruct 3D human mesh
using the predicted SMPL parameters.
Figure~\ref{fig:supp_qualitative_smpl} shows the visualization of the estimation results obtained by our FastMETRO and the SMPL parameter regressor.

\qquad \qquad 

\end{document}